# Network Engineering for Complex Belief Networks

**Suzanne M. Mahoney**
IET, Inc.
3710 N. Lynn
Arlington, VA 22009
suzanne@iet.com

**Kathryn B. Laskey**
Dept. of Systems Engineering and C³I Center
George Mason University
Fairfax, VA 22032
klaskey@gmu.edu

## Abstract

Developing a large belief network, like any large system, requires systems engineering to manage the design and construction process. We propose that network engineering follow a rapid prototyping approach to network construction. We describe criteria for identifying network modules and the use of 'stubs' within a belief network. We propose an object oriented representation for belief networks which captures the semantic as well as representational knowledge embedded in the variables, their values and their parameters. Methods for evaluating complex networks are described. Throughout the discussion, tools which support the engineering of large belief networks are identified.

## 1. INTRODUCTION

As belief networks become more popular and well understood as a tool for modeling uncertainty and as the computational power of belief network inference engines increases, belief networks are being applied to problems of increasing size and complexity. In the early 1990's, Pathfinder, at 109 nodes, was considered a large belief network (Heckerman, 1990). Pradhan et al. (1994, p. 484) reported that the 448 node CPCS-BN network was "... one of the largest BNs in use at the current time ...". But applications are rapidly pushing beyond these limits. As members of an IET knowledge elicitation and tool development team, we have been building a collection of coupled networks, each of which models a component or aspect of a military situation awareness problem. Most of these networks contain a hundred or more nodes, and the total number of variables for the completed system will easily number in the thousands. IET's PRIDE© software is being used to construct and manage these networks.

While the complexity of problems tackled by practitioners has exploded, the literature on knowledge engineering has not kept pace. There is a fairly extensive literature on probability elicitation dating from the 1970's (e.g., Spaetzler and Stael von Holstein, 1975; see Morgan et al., 1990 for an overview and some extensions). Much of this literature focused on how to elicit a probability distribution for a single variable, with some discussion of the need to "extend the conversation" to include other variables on which the distribution depended. The ability to build complex multivariate models was limited at that time by the available modeling technology and computing capability. With the advent of influence diagrams (Howard and Matheson, 1981) and Bayesian networks (Pearl, 1988), the complexity of problems to which probabilistic methods could be applied took a quantum leap. However, there remains a dearth of literature on knowledge engineering methodologies to help applications practitioners put the new technology to use. Pearl (1988) includes some helpful discussion on the qualitative meaning of the independence assumptions in a Bayesian network, but the primary emphasis of the book is on algorithms. Heckerman (1990) develops methods to modularize and improve the efficiency of knowledge elicitation for diagnosis networks with a single hypothesis variable. Pradhan et al (1994) and Provan (1995) discuss modeling approaches and computer support for diagnosis networks with multiple hypothesis nodes; the approach makes heavy use of the properties of the leaky noisy-OR. But there is as yet no source that an applications team can consult to help structure the difficult process of building a belief network model for a complex problem.

Sound network engineering means application of the systems engineering process to the design of complex belief network models. Section 2 of the paper describes the systems engineering process and its application to the design of belief network systems. The remainder of the paper discusses a set of issues that are key to the network engineering process. These issues are:
(1) *Modularization.* Decomposition into separable subproblems is at the heart of systems engineering.
(2) *Object-orientation.* The object-oriented paradigm provides a set of concepts and tools that simplify construction and maintenance of complex models.
(3) *Knowledge bases.* The knowledge that goes into a complex belief network model is far more than a set of nodes, states, arcs and probability distributions.
(4) *Evaluation.* Development of a complex network model requires constant evaluation during the

development cycle as well as testing of the completed model.

## 2. SYSTEMS ENGINEERING FOR COMPLEX BELIEF NETWORKS

A system is a set of interacting components organized to serve a common objective. Systems engineering is the process by which an operational need is transformed into a system that meets the operational need (Blanchard and Fabrycky, 1990; Sage, 1992). While the details of the process and the level of formality with which it is carried out are tailored to the particular requirements of a given application, the basic elements are common across applications.

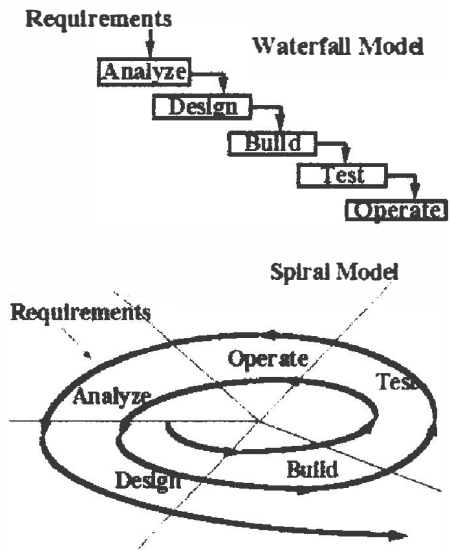

Figure 1. The Waterfall and Spiral Models

From initial conception through design, development, operation and finally, retirement, systems follow a predictable life cycle. The systems engineering process is organized around this life cycle, with activities that support the current phase and anticipate and plan for future phases. For our purposes there are two main categories of life cycle model: the waterfall and the spiral (Figure 1). In the waterfall model, system development is thought of as a linear step-by-step process of steps proceeding from design to development to operation. Waterfall models allow feedback between steps, but these are thought of as error correction mechanisms. The spiral model (Hall, 1969; Boehm, 1988) views system development as a repeating cycle of design, development, operation and evaluation. Each evaluation phase is used to examine lessons learned and plan the next cycle of the development effort. The product developed in early cycles is typically called a *prototype*. Its purpose is learning and planning, not operation. At some point the prototype becomes a system *version*, and future cycles develop new versions of an operational system.

Knowledge engineering is best thought of as a process of discovery, not of extraction. For this reason, the spiral life cycle is most appropriate as a model of the network engineering process. As network construction progresses, the expert's and knowledge engineer's understanding of the problem deepens. Exploring a prototype network's behavior on even a highly simplified problem fills in voids in the knowledge engineer's understanding of the domain and the expert understanding of how a belief network "thinks" about the problem. Managing this evolutionary process is the task of network engineering.

As with other engineering domains, network engineering begins with a set of task requirements. This is true whether the belief network stands alone or is embedded within a larger application. In our military domain, the system is supposed to answer the question: given observations of a possible unit at a specified location and time, what type of unit is present, how long will it be there and where is it likely to be next? The equipment on the ground and the ground itself are the features which are usually observed. The actual observations come in many forms: imagery, communications intercepts, signals of various types, terrain data bases and even human observers. The task is further complicated by having to deal with any number of units at many possible locations across a battlefield over time.

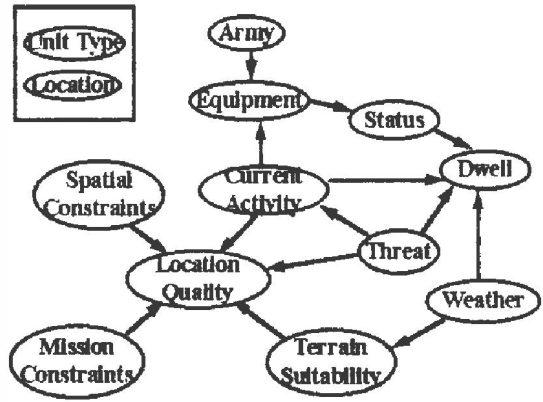

Figure 2. Activity Model

A major network engineering issue is to select a series of prototype models. An initial model should be reasonably small and self-contained so that it can be ready in a relatively short time. We chose to limit our initial prototype to a snapshot in time for a relatively small number of units and only include features that can be observed or inferred and not the observations themselves. We chose to build prototype activity models for a subset of unit types. Figure 2 shows a highly abstracted version of the activity models. The box in the upper right hand corner shows the conditioning variables for the model. The actual models contain about 50 variables each for each unit type modeled. Each model assumes a given unit type at a specified location. The structure of the problem is similar across unit types. We applied the lessons learned from modeling one unit to modeling the others. There should be a natural progression through a sequence of prototypes to completion of the full model. In our case, our initial models were expanded to include observations by a selected subset of sensors. Directions in which to expand next include addition of more unit types, addition of more sensor types, the task of discriminating among different unit types, the task of inferring a unit's location, and the task of aggregating units into higher level units.

## 3. MODULAR DECOMPOSITION

The first step in designing any complex system is decomposition into separable subproblems. For belief network models, decomposition is necessary for computational tractability, for comprehensibility by both modeler and expert, and for feasibility of testing the model. Belief networks were developed as a tool for modular decomposition of multivariate probability distributions. The modules in a belief network are the *local distributions*. Each local distribution consists of a variable, its parents, and a set of conditional distributions for the node given each combination of values for its parents. As a problem becomes more complex, a single level of decomposition becomes insufficient. When there are hundreds or thousands of variables, the entire network is far too complex to comprehend as an entity, while the local distributions are at much too low a level to serve as basic components of design. Design of a complex belief network system requires working at an intermediate level, in which the problem is decomposed into a set of coupled subnetworks, each of which represents a partially separable component of the problem.

For example, similarity networks (Heckerman, 1990) were designed as a tool to allow an expert to make probability assessments for a subproblem of a complex diagnosis problem. In the Pathfinder medical diagnosis problem, the expert restricted attention to a small set of diseases and to findings relevant to distinguishing among these diseases. By repeating this process for overlapping subsets, the expert implicitly specifies a global belief network for the entire diagnosis problem.

Although similarity networks are relevant to some aspects of the military situation awareness problem, the required assumptions are not met for other major subproblems. Nevertheless, the basic approach of decomposing into loosely coupled components is still necessary. Components in such a decomposition must be both *semantically separable* and *formally separable*. Semantic separability means that the subproblems into which the problem is decomposed are meaningful to the expert and posed at a natural level of detail. Formal separability means that the subproblems are capable of being reaggregated into a complete and consistent probability model. Both local models in a belief network and the local networks in a similarity network are formally separable. Multiply sectioned Bayesian networks (MSBNs) are another representation in which a network is disaggregated into formally separable components (Xiang, et al, 1992). That is, interfaces between components of a MSBN are required to satisfy conditions that ensure that all components taken together form a complete and consistent probability model.

In our military situation domain, each unit model is composed of three components: activity cycle; location quality; and equipment. We worked these out separately before combining them into a single unit model because it was easier to comprehend each subproblem separately. Other submodels which were not as relevant to the central activity model for the unit were represented by stubs (see below). The unit models themselves may be aggregated into higher level unit models (e.g. batteries within a battalion). They are also components of an identification model for discriminating among the different types of units. In the case of higher level unit models, several copies of a component may appear in a single model. Just maintaining consistency for a single component across multiple models is a difficult task. We are working toward a model library for network construction, maintenance and configuration management (Section 5).

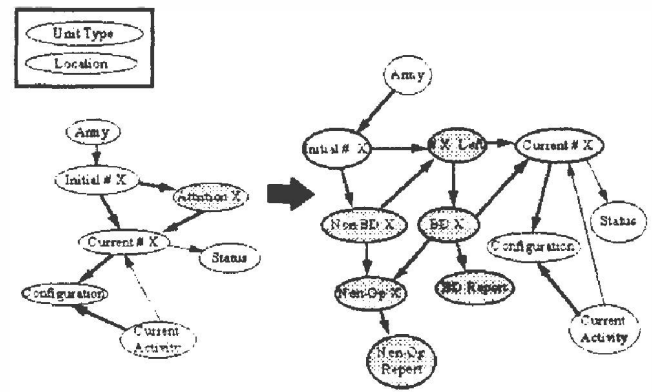

Figure 3. Stub and Model for Attrition

Sometimes it is not possible or desirable to focus on an entirely separable problem for a prototype model. In these cases, non-prototyped elements may be represented by stubs. A stub for a network component may be represented by one or more variables. For example, in our military situation model we modeled the output from a terrain database with a simple two state variable which represents whether or not the terrain at the location under consideration is suitable for a specified activity. In another case, we generated notional numbers for a greatly simplified attrition model. In both cases the stub served to model the interface between the stubbed component and the activity model we were prototyping. Figure 3 shows the stub for the attrition model as it was initially implemented and the more elaborate model that was developed when observations were added to the prototype. These and other network stubs allow the activity model to be tested without requiring the internal structure and parameters of the stubbed models to be fully elaborated.

The knowledge engineer works with the expert to decompose the problem into semantically separable components at an appropriate level of aggregation, and to ensure that these components are also formally separable so that they can be composed into a global model. Although existing representations such as similarity networks and MSBNs provide a set of tools for knowledge engineers to work with, the current toolkit is not sufficient for all problems. Additional research is required to develop representations for building formally separable submodels of complex problems. For example, the features of a 'good' Bayesian network module need to be worked out. A second key task of the knowledge engineer is to work locally within submodels to build models for subproblems. A third key task is to maintain interfaces between submodels to permit eventual integration of the modules. Representation frameworks and data management strategies (Section 5) that facilitate these tasks are important to effective knowledge engineering.

## 4. OBJECT ORIENTATION

The object oriented way of thinking has made rapid inroads in the software engineering community because it offers a set of tools and concepts that simplify construction and maintenance of complex software systems. Key concepts of the object oriented approach include (Booch, 1991):

• *Abstraction*. The designer identifies and abstracts "the essential characteristics of an object that distinguish it from ...other...objects and thus provide crisply defined conceptual boundaries..." (Booch, 1991, p. 39)

• *Encapsulation*. Details about the object that "do not contribute to its essential characteristics" (Booch, 1991, p.46) are hidden from the external world.

• *Hierarchy*. Abstractions are ordered hierarchically. Characteristics of an abstract class need not be specified directly for particular subclasses, but can be inherited from the abstract class.

These concepts of abstraction, encapsulation and hierarchy can be exploited to streamline knowledge engineering, to reduce data entry, and to simplify maintenance of a complex knowledge base. Our object-oriented approach is a natural extension of the frame-based (Goldman, 1990) and rule-based (Breese, 1990) approaches, and extends their representational power to include methods as well as data for belief network elements.

Some abstractions may apply to classes of variables. Figure 4 shows part of a class hierarchy for the many variables which represent distances. The is-a hierarchy is shown on the left with examples of inheritable features shown on the right.

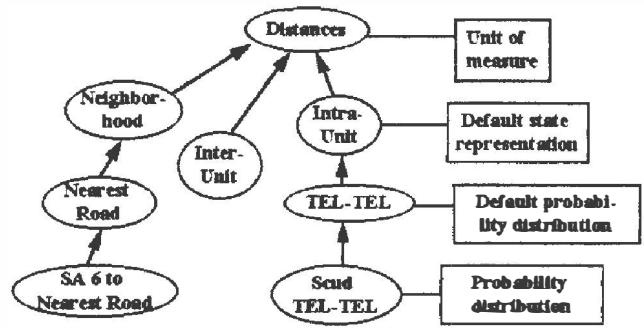

Figure 4. Class Hierarchy for Distance

While class hierarchies for single variables provide consistency across models and reduce data entry, their usage is limited compared to that of repeatable structural features within the domain. These repeatable structures can be abstracted as objects, with encapsulated internal structure, and organized hierarchically. For example, Figure 5 shows a progression of models for dwell. Note that the initial General Dwell Model is a subset of the variables in Figure 2. The Missile Unit and Surface to Surface Missile Dwell Models each have more variables and relationships than the more abstract model. We have exploited the commonalities represented in the abstractions across different types of missile units to facilitate our knowledge engineering. The lessons which we have learned from doing SCUDs and SA6s in depth are now being transferred to the other missile unit types. Both we and the domain experts are becoming quite proficient at building new models around a familiar structural template. As objects these models may include methods as well as states and probability distributions. Methods can be used to reduce knowledge elicitation and data entry burdens and to capture relevant knowledge which is not representable in a Bayesian network. For example, we anticipate incorporating a method for simulating Dwell into a Missile Unit Dwell Model.

Templates in the form of parameterized abstractions are also present in our complex belief network environment. An example of a template model is shown in Figure 3. The attrition model template's parameters are the equipment type, replacing X, and a range of numbers for the states of Initial # X. This template can be reused across all equipment types and greatly reduces the effort involved in generating new models.

We have been discussing the application of object-oriented ideas to repeated structural features within the domain. There are also repeated structures that occur across domains, that are inherent to the network engineering process. One example is the existence of standard distribution types. These include the noisy-OR (Pearl, 1988), the normal distribution, and distributions defined over partitions of the parent variable state space (Heckerman, 1990). Another example is the definition of operations that occur commonly in belief network construction. These include discretization of continuous variables, decision analytic methods for reasoning about the level of abstraction of a state space (Poh, 1993), or methods for extracting a subnetwork for which inference results approximate or equal those for the full network (see Breese, 1990; Provan et al., 1994). Again, these repeated structures and processes represent opportunities for abstraction and encapsulation.

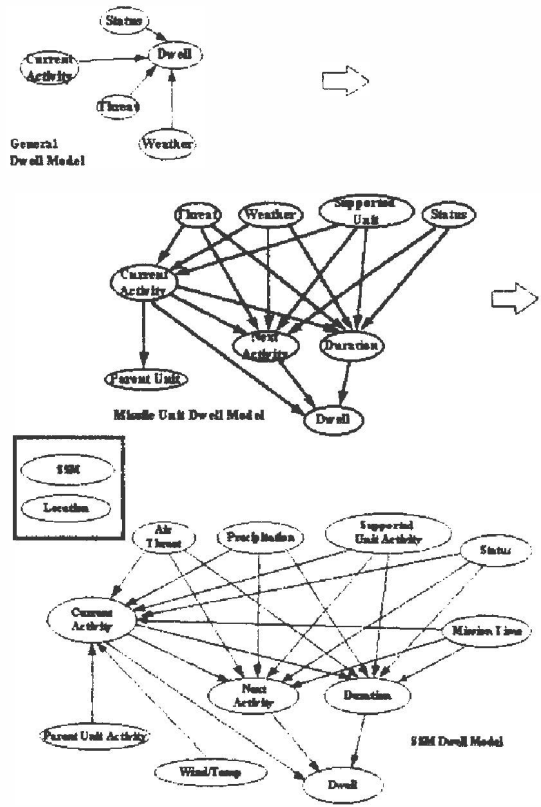

Figure 5. A Progression of Dwell Models

## 5. REQUIREMENTS FOR A KNOWLEDGE BASE

Object oriented analysis concentrates on design implications for integrating elements of the fundamental Bayesian network knowledge representation with the domain's concepts. Knowledge base support concentrates on the structure and contents of the knowledge base which is required to support the knowledge engineering process. While there has been some research on representation (Heckerman, 1990; Poh, 1993), it has lagged behind progress in inference and has not been implemented in the most widely available tools. Thus, most knowledge engineers find themselves working with representation technology that supports inference far better than it supports the knowledge engineering process.

Too much of the knowledge which goes into creating a Bayesian network model cannot be represented explicitly using current knowledge representation methodology and current software tools.

The fundamental requirement for a knowledge representation framework is the ability to capture and maintain knowledge at the semantic level. Neither the expert nor the knowledge engineer should be required to work directly at the inference engine level. The translation from semantic level to inference engine level should be automated as much as possible. Bayesian networks have been promoted as freeing the developer to think about structure (i.e., conditional independencies in the domain) rather than the details of inference. But as the problems tackled become more complex, richer kinds of structure become apparent and it becomes necessary to capture these structures and the semantic reasoning behind them in a knowledge base.

As one example, it has already been noted in the literature that knowledge elicitation can be simplified by taking account of subset independence (Geiger and Heckerman, 1991). Subset independence is supported by basing elicitation on *partitions* (Heckerman, 1990) of parent variable state spaces. If the child variable conditional distributions satisfy subset independence for each partition element, then only one distribution need be assessed for each partition element.

A corollary of the requirement to capture semantic knowledge is the need to provide explicit representation for semantic structure and not just the data derived from it. For example, one can represent subset independence in a standard Bayesian network simply by entering the same distribution for several parent state combinations. However, this strategy provides no way to encode the reason for the equality or distinguish it from an accidental equality. Moreover, the knowledge base is difficult to maintain. This general principle also applies in more complex cases. Examples include inequality constraints among distributions (cloud cover reduces the diagnosticity of optical imaging sensors; the probability of detecting an object decreases as the distance from the target increases) and structural commonalities among network fragments (weather and distance from target affect probability of detection for all imaging sensors). The ability to capture and enforce structural and numerical constraints provides important support to the expert in constructing a consistent model that truly captures his/her expertise. It also supports maintainability of the knowledge base.

The next iteration for military situation assessment requires model 'agility'. This means that we have to be able to change models on the fly as the situation changes. We expect to generate revised models through a combination of learning, queries to the user, queries to databases and other techniques. Mature model libraries will provide us with variables and structure for whole subsections of a network so that we can quickly cobble together large complex networks. Representation of semantic constraints supports model agility because it enforces consistency and provides semantic-level knowledge to support adapting a component to a new application.

A final requirement for knowledge base development is automated support for configuration management. Configuration management includes support for maintaining a history of evolving versions of a knowledge base, provisions for documentation of changes and their rationale, protocols for making and logging modifications to the knowledge base, and easy comparison of similarities and differences between different versions of the knowledge base.

## 6. EVALUATION

The success of the prototyping approach depends critically on evaluation. Evaluation provides information for prioritizing enhancements in the next iteration. It is important to stress that the purpose of evaluation should *not* be to showcase the system or demonstrate that it works. Evaluation should explicitly attempt to push the boundaries, to identify problems on which the system fails and areas in which improvement is needed.

The literature on evaluation distinguishes between verification and validation. (Adelman, 1992) Verification is concerned with measuring the degree to which a system meets the specifications for which it is designed. Verification of belief network models includes evaluating factors such as correctness of algorithms, functional completeness of the knowledge base, speed of processing, and satisfaction of interface requirements with other systems. Verification should also include checking the extent to which the design, coding and documentation of the system meet organizational standards. Validation measures the extent to which the system meets the operational need for which it was designed. However, at the current stage of our own knowledge base development effort, such testing is not yet feasible, and our writings on the subject would not be sufficiently informed by experience.

We performed three basic types of evaluation. The first type, which we call *elicitation review*, involves an overall review of node definitions, state definitions, independence assumptions, and probability distributions. Elicitation review permits the expert to take a more global view of a network module, examine the consistency of definitions and judgments, and adjust the model to achieve consistency. Because we were dealing with multiple domain experts this step was particularly critical to achieving consistency among the models. In particular the importance of maintaining a central repository for definitions became evident.

The second type of evaluation we call *importance analysis*. It is a form of sensitivity analysis (Laskey, 1993; Morgan et al., 1990) and is related to methods for explaining the output of belief network models (Suermondt, 1992). Importance analysis for a given variable (called the *focus* variable) measures the impact on the focus variable's belief of obtaining evidence about each of a set of other variables (the *evidence* variables). The method we implemented analyzes the evidence variables one at a time, measuring changes in the quadratic score of the focus variable (Weiss, 1995). The output of importance analysis is a plot for each focus variable that orders the evidence variables by their impact on the focus variable. (See Figure 6.) Experts reviewed these plots with elicitors. The models were updated to correct the inconsistencies detected by the experts. Synergistic effects of changes in multiple variables could be examined by statistical sampling of combinations of evidence variables (Morgan et al., 1990)

```
Importance Analysis for "Dwell"

Current Observations:     Country/Army: Iraq

IMPORTANCE  ##      NAME------------------------
            ----------------------
    ++++++++++  36 SCUD Battery Current Activity
Type
        *****   20 Supported Unit Activity
        ****    14 Location Mission
Supportability-- Distance to Target
        *        5 Parent Unit Activity
        *        5 Report # Washdown Vehicles
        *        3 Precipitation
        *        2 Mission Time
        *        2 TEL-TEL Dispersion
                 2 Air Threat
                 1 Location Accessibility
                 1 Wind/Temp
                 1 SOF Threat
                0+ Report # Survey Veh.
                0+ Status
                0+ Artillery Threat
                0+ Report # TELs
```

Figure 6. An Importance Analysis for Dwell for SCUD Batteries

The third form of evaluation was *case-based evaluation*. We worked with the experts to develop a set of *scenarios* for testing. Each scenario concentrated on one or two focus variables and a small set of evidence variables. A test instance consisted of an assignment of a state to each evidence variable. The experts reviewed the posterior distributions on the focus variables for each test case. In applications in which outcome data are available, posterior beliefs can be compared with "ground truth" values for the tested scenarios. The case-based evaluation was supported by a spreadsheet which generated the test case specifications from parameters. These documented tests will become a valuable source for regression tests as the system evolves. We are working towards further automating the case based testing process. Automation is particularly important for regression testing.

The literature on software testing stresses *coverage*. In our problem, coverage corresponds to the proportion of possible test instances. For relatively small scenarios it is possible to achieve high coverage. Test cases should cover typical, infrequent, and unanticipated conditions. Experts may have difficulty evaluating the behavior of the system on unusual cases. Some of these may be classified as outside the scope of the model. The completed system should include a filter for identifying and alerting users to out-of-scope scenarios (see Laskey, 1991; Jensen, et al., 1990 for approaches to diagnosing out-of-scope scenarios).

When moving beyond the module level to the system level, the biggest difficulty that arises is sheer combinatorics. If modules are designed well, this difficulty should be mitigated due to a high degree of conditional independence between modules. However, this assumption must be verified.

## 7. SUMMARY

As models become more complex and greater agility is required, the discipline of systems engineering becomes necessary to network construction. This paper discusses a number of important issues in the application of systems engineering to belief network development. The key point is that like any complex entity, networks have to be engineered. Systems engineering defines the important elements of any construction process. What those of us working with belief networks must provide are the specific methods and tools to support systems engineering for complex networks.

Modularity in a network is key to the decomposition process that accompanies systems engineering. Achieving formally separable model components which are semantically meaningful is a difficult task. More research is needed to provide generic structure for the knowledge engineer's toolkit.

The object oriented paradigm provides the community with a means to parsimoniously represent data and knowledge associated with Bayesian networks. We have described abstractions, method encapsulations and hierarchies suggested by our work which show that knowledge engineers can benefit significantly by thinking beyond the context of individual variables and their parameters.

The knowledge incorporated into a Bayesian network goes well beyond the names and numbers that are usually represented. There are relationships among the numbers and among the variables and their states which generally go unrepresented. These qualitative constraints enforce consistency within a model. Representing and enforcing them will be critically important as networks are required to become more agile. Just as important are tools including libraries and configuration management tools which support the knowledge as it evolves.

Evaluation also has to be rethought in the light of complex networks. Testing simply cannot be done only at the network level. It becomes an ongoing process as the network is developed. Like the knowledge, evaluation for networks needs to be supported by an appropriate tool set.

## Acknowledgments

We particularly want to thank Tod Levitt, Anne Martin, and Jonathan Weiss of IET, and the domain analysts at Pacific-Sierra Research Corp. for their work towards developing the models described in this paper. We are especially thankful to K-C Ng of IET for his work on PRIDE®. The work on the military situation modeling was funded by DARPA under contract number DACA 76-93-C-0025. We also thank the anonymous reviewers for their insightful comments.